\pdfoutput=1

\documentclass[11pt]{article}

\usepackage{ACL2023}

\usepackage{times}
\usepackage{latexsym}

\usepackage[T1]{fontenc}

\usepackage[utf8]{inputenc}

\usepackage{microtype}

\usepackage{inconsolata}

\usepackage{graphicx}
\usepackage{amsmath}
\DeclareMathOperator*{\argmaxA}{arg\,max}

\title{AdamR at SemEval-2023 Task 10: Solving the Class Imbalance Problem in Sexism Detection with Ensemble Learning}

\author{Adam Rydelek, Daryna Dementieva, \and Georg Groh \\
  TU Munich, Department of Informatics, Germany \\
  \texttt{\{name.surname\}@tum.de} \\ \texttt{grohg@in.tum.de}}

\begin{document}
\maketitle
\begin{abstract}
The Explainable Detection of Online Sexism task presents the problem of explainable sexism detection through fine-grained categorisation of sexist cases with three subtasks. Our team experimented with different ways to combat class imbalance throughout the tasks using data augmentation and loss alteration techniques. We tackled the challenge by utilising ensembles of Transformer models trained on different datasets, which are tested to find the balance between performance and interpretability. This solution ranked us in the top 40\% of teams for each of the tracks.
\end{abstract}

\section{Introduction}

Sexism and other forms of harassment can be commonly observed on the internet. It can not only discourage attacked groups from interacting on the internet but also pose a serious threat \cite{fox15}. Especially with the growing popularity of social media platforms, the need for automation in detecting such occurrences arises. Such tools already exist and are being constantly developed but still face multiple issues \cite{Gongane2022}. One of the present problems in such systems is often the lack of reasoning and interpretation behind the prediction itself. Presenting an explanation why the system has predicted a text to be sexist enhances trust in such systems and enables them to be more widely used \cite{ferrario22}. The Explainable Detection of Online Sexism (EDOS) task presents a problem of detecting sexism in English texts gathered from popular Social Media platforms: Reddit and Gab \cite{edos2023semeval}. The detection problem is proceeded by two subtasks: firstly, classifying the sexist entries into one of four categories, and finally, marking it as one of the eleven fine-grained vectors.

In this work, we propose systems to tackle each of the subtasks which share the architecture and data pre-processing methods. For each problem, an ensemble of Transformer-based models is built and fine-tuned on English texts concerning the subtask. Models within the ensemble are trained on different sets of data. The provided EDOS datasets have been extended by data points from various sources concerning past research on the topic. The texts have been pre-processed and normalized to fit the schema provided with the train data to enhance the results. The performance metric chosen for the task is the macro-averaged F1 score, which treats each class equally and promotes balanced classification.

Our contributions consist of providing ensemble-based solutions to all subtasks ranking us in the top 40\% for each of them---36th place for Sexism Detection, 29th for Category of Sexism, and 25th for Fine-grained Vector of Sexism tracks. We also test different methods of data augmentation and architecture designs to tackle class imbalance problems. A comparison of different ensembles is presented, weighting its advantages and disadvantages over using a single model.

\section{Background}
\label{sec:background}
The task of explainable sexism detection consists of three subtasks. Task A is the binary classification of sexism. Task B considers only sexist texts and classifies them into one of four distinct categories: threats, derogation, animosity, and prejudiced discussions. Task C breaks down the categories into an even more fine-grained vector of eleven possible classes. The provided training dataset consists of 14000 data points, each containing an English text and three labels for each subtask. The texts have been scraped from Gab and Reddit in equal proportions and labelled by three annotators with supervision from two domain experts. Classes for task A are imbalanced and skewed towards the `non-sexist' class, with 10602 entries. This translates to just 24.3\% of labels belonging to the minority class. A similar challenge is present in both task B---the smallest class contains just over 9\% of entries and task C---only 1.4\% of entries belong to the `condescending explanations or unwelcome advice' class. The texts have already been pre-processed in some regard, as the usernames and URLs have been replaced with a special token. 

The problem of automated sexism detection in Social Media has already been tackled by multiple researchers in the past years as it gained traction as a crucial issue along with the development of Natural Language Processing (NLP) techniques allowing to overcome it more efficiently. SEXism Identification in Social neTworks (EXIST) is a shared task focused on sexism detection proposed by the organisers for the IberLEF 2021 campaign \cite{Rodrigues21}. With the first edition's success, it was followed by a similar challenge in the following year which provided an overview of leading techniques for tackling such problems \cite{Rodrigues22}. The relevance of those tasks is even more considerable for the problem presented in EDOS, as EXIST also introduced a more fine-grained categorisation of sexism into five categories. The leading solutions for the latest instalment of the challenge were based on the Transformer architecture, a deep neural network architecture based exclusively on attention \cite{transformers17}. The authors concluded that Transformer models adapted to the social media domain improve the accuracy of sexism detection. The best-performing models proved to be majority-vote ensembles of multiple models with varying hyperparameters. Superior results for the task were achieved using only the text embeddings without additional features. However, Waseem and Hovy showed on a similar task of hate speech detection on Twitter that demographic information can provide an improvement on the model \cite{waseem-hovy-2016-hateful}.

\section{System overview}

The system proposed in this work to tackle each task consists of five mutual steps: data collection, data analysis and preprocessing, model training and development, hyperparameter tuning and model selection, and final ensemble implementation. While the utilized data, models, and implementation varies between subtasks, the methodology remains the same. Starting from data collection, where additional fitting data is obtained apart from the provided resources, which is then analyzed, processed and normalized. On the combined data, multiple models with different architectures are trained or fine-tuned and undergo hyperparameter tuning on the validation dataset. Based on the results on selected test sets, a subset of models is selected, and an ensemble is created.

\subsection{Data collection}

Multiple additional datasets have been used concerning similar or identical tasks to extend the available training data and enhance predictive capabilities. The EXIST task mentioned in the previous section introduced a high-quality dataset on sexism detection \cite{Rodrigues21}. It consists of 11345 data points, each containing the source of the texts---Gab or Twitter, the language of the text - English or Spanish, the text itself, along with the binary sexism label and a more fine-grained category of sexism. It was labelled by five crowdsourcing annotators on data scraped from those social media platforms. English texts amount to just under 50\% of texts with 5644 entries. Contrary to the EDOS dataset, EXIST data has been balanced to a more even split between the binary classes, with 49.5\% of the entries being sexist. It has not been preprocessed and contains raw, scraped texts. Another used source of additional training data was the HatEval dataset introduced for the 2019 SemEval competition \cite{hateval19}. It consists of 19600 Tweets concerning the topic of hate speech detection targeted at two specific groups: women and immigrants. Similarly to the EXIST data, HatEval texts consist of Spanish and English entries, with the latter amounting to 13000. For this task, 5910 entries were chosen from this dataset, of which 2510 were labelled as hateful towards women. Multitarget-CONAN was another used dataset that focuses on hate speech and counter-narrative pairs that threaten multiple target groups \cite{chung-etal-2019-conan}. It consists of 5003 entries targeting seven different groups, from which 662 hateful texts towards the women group have been selected to be added for this task. The final dataset consists of 28873 data points, of which 10059 are sexist, doubling the number of instances from the provided EDOS training data.

For experiment purposes, an even more extensive dataset has been created by inserting the Spanish entries from both the EXIST and HatEval datasets, along with the MeTwo dataset, which provides Twitter sexism detection data in Spanish \cite{metwo20}. This allowed to include 14134 additional entries that have been translated using the Google Translate API - googletrans\footnote{\href{https://py-googletrans.readthedocs.io/en/latest/}{https://py-googletrans.readthedocs.io/en/latest/}}.

\subsection{Training data preprocessing}
\label{sec:eda}

The provided dataset from the SemEval 2023 task has already been preprocessed, introducing a special token for usernames and URLs. For consistency reasons, the whole dataset has been tokenized in this way. Along with that, all e-mail addresses, phone numbers, and currency symbols have been replaced with according tokens. All of the non-ASCII symbols have been transliterated to the closest representation. Punctuation and casing have been left unmodified after prior testing. Texts from the Multitarget-CONAN dataset had the word `woman' or other variation of it in more than half of the sexist entries; hence they were replaced with synonyms to avoid overfitting to this specific word. This allowed for a concise and coherent dataset ready for further development.

\begin{figure}[ht!]
\includegraphics[width=\columnwidth]{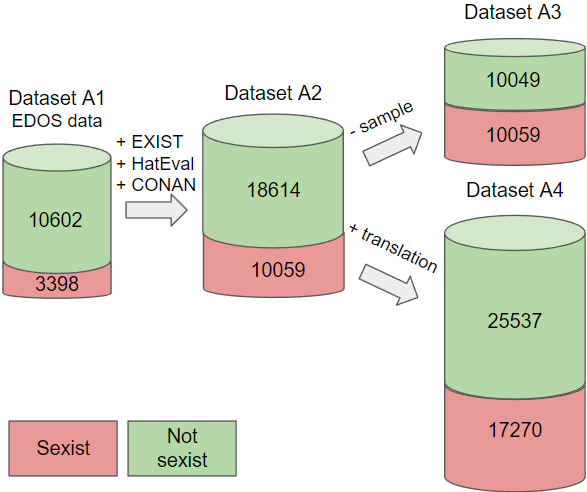}
\caption{Illustration of datasets created for task A. The number of samples is defined inside cylinders.}
\label{fig:dataset}
\end{figure}

For Task A, the whole dataset was split into four subsets presented in Figure \ref{fig:dataset}: dataset \textbf{A1} - the base dataset provided by the SemEval 2023 task organisers, dataset \textbf{A2} - all the entries natively English, dataset \textbf{A3} - a balanced version of dataset A2 to achieve 50/50 target class split, dataset \textbf{A4} - all the available data with translated texts from Spanish. For the balanced version, the entries removed to allow for an even class balance were taken from the additional datasets to retain all of the provided training information. 

\begin{figure}[ht!]
\includegraphics[width=\columnwidth]{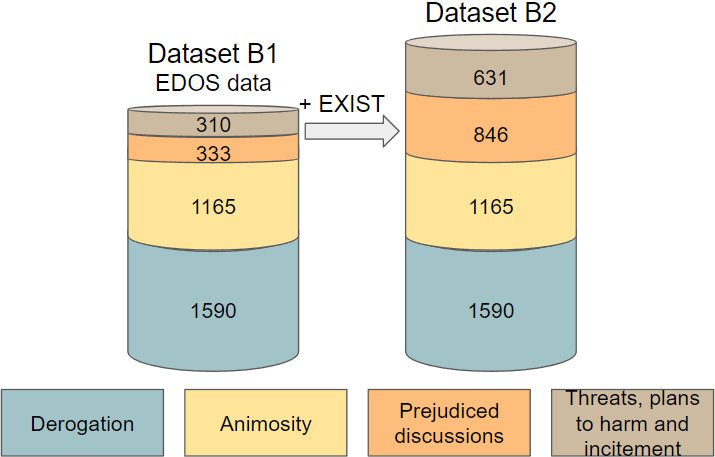}
\caption{Illustration of datasets created for task B. The number of samples is defined inside cylinders.}
\label{fig:datasetB}
\end{figure}

Task B introduced a multi-label classification problem for sexist entries. This meant that the provided training data consisted of just 3398 sexist entries for this subtask, which will be called dataset \textbf{B1}. As most of the external data sources consisted only of binary sexism detection labels, they were purposeless for this problem. The only dataset containing a further fine-grained classification of sexist cases was EXIST. The problem with EDOS training data was the vast imbalance in the distribution of the four present classes. The `derogation' class accounted for 46.8\% of the data points, `animosity' for 34.3\%, while `prejudiced discussions' only for 9.8\%, and `threats, plans to harm and incitement' for 9.1\%. Additional data from EXIST was introduced to remedy the lack of entries in two minority classes. A RoBERTa Large sentence Transformer model has been used to obtain the embeddings, allowing us to calculate distances between texts from the two datasets \cite{reimers-2019-sentence-bert}. For all points in the minority class of `threats', the texts have been embedded using the model. Similarly, all English texts from EXIST data have been embedded. For each sample in EXIST, Cosine similarity to all embeddings from the `threats' class was calculated and averaged for a single score. Cosine similarity is a common metric for assessing similarity between two vectors and is calculated as a dot-product of normalized vectors \cite{baoli13}. For n-dimensional vectors \(\vec{v}\) and \(\vec{w}\), the Cosine similarity is defined by the expression:
\begin{equation}\label{eq:sim}
\begin{aligned}
    Sim(\vec{v},\vec{w}) = \frac{\vec{v} \cdot \vec{w}}{| \vec{v} | | \vec{w} |} = \frac{\sum_{i=1}^{n}v_i w_i}{\sqrt{\sum_{i=1}^{n} v_i^2} \sqrt{\sum_{i=1}^{n} w_i^2}}
\end{aligned}
\end{equation}
A threshold of 0.45 similarity has been chosen after testing, and entries from EXIST that had an average Cosine similarity under it were filtered out. In this subset, there were primarily texts from two categories: `sexual-violence' and `misogyny-non-sexual-violence'; hence, they were chosen and inserted as `threats, plans to harm and incitement' into the training data. A similar procedure has been conducted for the `prejudiced discussions' class, which resulted in additional entries from classes `ideological-inequality' and `stereotyping-dominance'. The resulting dataset presented in Figure \ref{fig:datasetB} is more balanced, and the contribution of minority classes rose from 9.1\% to 14.9\% for `threats, plans to harm and incitement' and from 9.8\% to 20\% for `prejudiced discussions'. The extended dataset consists of 4232 entries and will be mentioned as dataset \textbf{B2} in further development.

\subsection{Model development}

\paragraph{Baseline} The development started with baseline experiments to facilitate an entry point for Task A---binary classification. TF–IDF (term frequency–inverse document frequency) is a weighting strategy used to represent text data as vectors, which is commonly used \cite{ref1}. The experiments started by training SVM and Naive-Bayes classifiers on the base dataset A1 \cite{cortes1995support}.

\paragraph{CNN} Further development contained training Convolutional Neural Networks on the A1 dataset \cite{cnn15}. Texts were represented using the base GloVe (Global Vectors for Word Representation) embedding \cite{pennington-etal-2014-glove}. The architecture of the network consisted of three convolutional layers, which were optimized using the Adam optimizer with default values \cite{adam14}. It was meant to serve as a baseline value; hence the architecture was not further optimized for the task.

\paragraph{Transformers} Basing on previous research mentioned in Section \ref{sec:background}, the primary focus was put on testing Transformer models. The first architecture selected for testing was the commonly used BERT model family \cite{devlin-etal-2019-bert}. Along with the base, uncased version of BERT, BERTweet was also used as it was pre-trained on 850 million English Tweets, fitting the task at hand \cite{nguyen-etal-2020-bertweet}. Base and large versions of RoBERTa were also used, which provides an optimized version of BERT pretraining \cite{roberta19}. A RoBERTa model\footnote{\href{https://huggingface.co/s-nlp/roberta_toxicity_classifier}{https://huggingface.co/s-nlp/roberta\_toxicity\_classifier}} introduced by Skolkovo Institute that was already fine-tuned on the Jigsaw toxic comment datasets\footnote{\href{https://www.kaggle.com/c/jigsaw-multilingual-toxic-comment-classification}{https://www.kaggle.com/c/jigsaw-multilingual-toxic-comment-classification}} has been selected to study the impact of pretraining on domain-specific data in comparison to the base model \cite{logacheva-etal-2022-paradetox}. Apart from that, XLNet was used as an example of a cased model in its base form \cite{xlnet19}.

\begin{table}[]
\begin{tabular}{l|l|l|}
Model             & Dataset & Macro-F1          \\ \hline
SVM (TF-IDF)      & A1      & 0.715             \\ \hline
CNN (GloVe)       & A1      & 0.754             \\ \hline
XLNet base cased  & A1      & 0.776             \\ \hline
BERT base uncased & A1      & 0.785             \\ \hline
RoBERTa base      & A1      & 0.791             \\ \hline
BERT base uncased & A2      & 0.795             \\ \hline
RoBERTa base      & A2      & 0.806             \\ \hline
BERTweet base     & A2      & 0.817             \\ \hline
RoBERTa toxicity  & A2      & 0.814             \\ \hline
BERT base uncased & A3      & 0.802             \\ \hline
RoBERTa base      & A3      & 0.810             \\ \hline
BERTweet base     & A3      & 0.816             \\ \hline
RoBERTa toxicity  & A3      & \textbf{0.820}    \\ \hline
BERT base uncased & A4      & 0.776             \\ \hline
RoBERTa base      & A4      & 0.782   
\end{tabular}
\caption{Table depicting the task A benchmark results of multiple models trained/fine-tuned on different datasets explained in Section \ref{sec:eda}. The Macro-F1 values have been calculated on the EDOS development set.}
\label{table:1}
\end{table}

\paragraph{Task A} Selected Transformer models have been fine-tuned on datasets A1, A2, A3, and A4 to select the most prominent ones for further optimization. Default parameters were used, meaning a batch size of 8, maximum length of 128, dropout equal to 0.3, and utilized the AdamW optimizer with Betas equal to 0.9, 0.999, and a starting learning rate of 5e-5 \cite{adamw17}. The training lasted for ten epochs, and the final models for the benchmark were chosen based on the validation set decided in an 80/20 split. Models were benchmarked on official EDOS development data provided by the task organizers, which have been depicted in Table \ref{table:1}. The results showed that Transformer models exceeded other methods as expected. The toxicity RoBERTa model and BERTweet base proved to show the best results and outperformed their competitors pre-trained on general data. Additional data provided in dataset A2 improved the results, but inserting translated data points from Spanish in dataset A4 resulted in significant downgrades in performance. The difference between models trained on the extended dataset A2 and its balanced version---A3----were minor in terms of Macro-F1, so both of them will be explored further to uncover the impact on performance on the `sexist' class.

\paragraph{Task B} The classification problem in task B requires multi-label classification while restricting the amount of available data drastically compared to the binary problem. With the results of task A in mind, baseline models were omitted during testing, and only Transformer models were taken into consideration. RoBERTa, BERT, and BERTweet have all been fine-tuned on both the SemEval dataset and the dataset extended with EXIST entries. All hyperparameters for this part of the testing were the same as in task A experiments. Results depicted in Table \ref{table:2} showed that BERTweet and RoBERTa achieved the best results and will be further used in ensemble creation. The insertion of EXIST data proved to decline performance in the case of RoBERTa models but slightly improved it when using BERTweet. 

\begin{table}[]
\begin{tabular}{l|l|l|}
Model             & Dataset & Macro-F1          \\ \hline
RoBERTa base      & B1      & \textbf{0.616}    \\ \hline
BERT base uncased & B1      & 0.589             \\ \hline
BERTweet base     & B1      & 0.600             \\ \hline
RoBERTa base      & B2      & 0.581             \\ \hline
BERTweet base     & B2      & 0.602   
\end{tabular}
\caption{Table showing the task B benchmark results of multiple models trained/fine-tuned on datasets B1 and B2 explained in Section \ref{sec:eda}. The Macro-F1 values have been calculated on the EDOS development set.}
\label{table:2}
\end{table}

\paragraph{Task C} The problem in task C is similar to task B in many ways but more fine-grained. Four classes present in task B are split into a vector of between 2 and 4 sub-classes for a total of 11 classes. Because of that, two approaches were tested---the first one being a single model for 11-class classification. The second approach was to use models derived from task B research to firstly label an entry as one of four task B categories, effectively splitting the dataset into four subsets. Based on that, a multi-class classification model was created for each subset. This approach allowed us to lower the number of possible classes and fight the vast imbalance problems. The latter strategy proved to be more efficient on the development set for all model architectures used, which were the same ones as in task B due to the similarity of the problem. As the problem of class imbalance is even more significant in task C compared to task B, a similar augmented dataset with EXIST data points was tested but proved to degrade the effectiveness of models both for BERT and RoBERTa.

\section{Experimental setup and final system implementation}

For final experiments and implementation, the datasets introduced in Section \ref{sec:eda} have been extended with provided development data (2000 data points) and pre-processed with the help of the clean-text\footnote{\href{https://pypi.org/project/clean-text/}{version 0.6.0 https://pypi.org/project/clean-text/}} package. After preliminary experiments have been completed, selected models have been retrained on the extended datasets using the Transformers\footnote{\href{https://pypi.org/project/transformers/}{version 4.26.1 https://pypi.org/project/transformers/}} package pipelines along with PyTorch\footnote{\href{https://pytorch.org/}{version 1.13.1 https://pytorch.org/}}. For validation purposes, 25\% of the entries in development datasets have been selected to achieve the same class split percentage as the primary dataset. For task A, it amounted to 500 validation data points; for tasks B and C, it was 100. The similarity for task B has been recalculated on the full dataset using methods described in Section \ref{sec:eda} and the sentence transformers\footnote{\href{https://pypi.org/project/sentence-transformers/}{version 2.2.2 https://www.sbert.net/}} package.

\subsection{Hyperparameter tuning and model selection}

Models selected in previous experiments underwent hyperparameter tuning with a focus on learning rate alteration. During a grid search run, the optimal value for BERTweet and RoBERTa base models appeared to be 5e-6, while for RoBERTa large 3e-6 was chosen. In hopes of achieving better performance in task A, all models fine-tuned on datasets A1 and A2, which are imbalanced, were tested using a weighted Binary Cross Entropy loss as it allows for harsher penalization of the model for getting the minority class wrong. For N training samples, the loss can be expressed as \ref{eq:weightedbce}, where the weight \(w\) is calculated as the inverse of the fraction of sexism samples to all samples, \(y_i\) is the i-th target label, \(h_\theta\) is the model with parameters \(\theta\), and \(x_i\) is the i-th input \cite{Ho_2020}.

\begin{equation}\label{eq:weightedbce}
\begin{aligned}
L_w = -\frac{1}{N}\sum_{i=1}^{N}(w y_i log(h_\theta(x_i))  \\
 + (1-y_i) log(1-h_\theta(x_i)))
\end{aligned}
\end{equation}

\paragraph{Task A} Using this modified loss function on dataset A1 without modifying remaining parameters increased BERT base performance from 0.785 to 0.789 and RoBERTa base from 0.791 to 0.8. On the other hand, it decreased performance for all models fine-tuned on dataset A2 and large model versions on the A1 dataset. The final selected models - RoBERTa large, BERTweet large fine-tuned on dataset A1, Toxicity RoBERTa, RoBERTa Large, BERTweet base on dataset A2, and Toxicity RoBERTa, BERTweet base on dataset A3 have been retrained using additional 2000 points from released development data, leaving out 500 mutual data points for validation of ensembles. The validation set has been selected to have a nearly identical distribution of sexist/non-sexist labels as the EDOS development set. Based on this set, the performance of tuned models has been measured, and the results can be observed in Table \ref{table:3}. 

\begin{table}[]
\begin{tabular}{l|l|l|}
Model            & Dataset & Macro-F1 \\ \hline
RoBERTa large    & A1      & 0.853    \\ \hline
BERTweet large   & A1      & 0.841    \\ \hline
Toxicity RoBERTa & A2      & 0.853    \\ \hline
RoBERTa large    & A2      & 0.861    \\ \hline
BERTweet base    & A2      & 0.851    \\ \hline
Toxicity RoBERTa & A3      & 0.847    \\ \hline
BERTweet base    & A3      & \textbf{0.865}   
\end{tabular}
\caption{Table showing task A final model benchmark. The Macro-F1 values have been calculated on a validation subset of 500 points. Dataset description is present in Section \ref{sec:eda}.}
\label{table:3}
\end{table}

While the Macro-F1 score does not show vast differences between augmented and base datasets, looking into separate metrics for each class allows us to see differences in operation. For the two models: RoBERTa large and BERTweet base, fine-tuned on each of the datasets, the averaged performance metrics for the `sexist' class were calculated and are depicted by Table \ref{table:test}. An enhancement in performance on the minority class can be observed in the case of the balanced dataset, as well as substantial differences in recall when compared to non-balanced data.

\begin{table}[]
\begin{tabular}{l|l|l|l}
Dataset & Precision          & Recall          & F1-score \\ \hline
A1      & 0.785              & 0.755           & 0.77     \\ \hline
A2      & \textbf{0.795}     & 0.765           & 0.78     \\ \hline
A3      & 0.775              & \textbf{0.79}   & \textbf{0.785}   
\end{tabular}
\caption{Table depicting the performance metrics for the `sexist' class, averaged from two models: RoBERTa large and BERTweet base fine-tuned on each dataset. Performance metrics have been calculated on a subset of 122 sexist samples from the validation set used in Table \ref{table:3}.}
\label{table:test}
\end{table}

\paragraph{Task B} For task B the group of models decided upon was BERTweet base, RoBERTa base, and RoBERTa large, which were trained on datasets B1 and B2. As the problem of vast class imbalance was still present in dataset B1, a different loss function was tested, allowing the model to focus on the most challenging entries. Focal loss modifies the weights for each sample by increasing them for the most difficult cases and reducing them for the easy ones \cite{focal2017}. For simplicity, in a binary case for \(y\) - ground truth label, \(p\) - estimated probability for label 1, \(\gamma\) - focusing parameter, \(\alpha\) balancing parameter, it can be expressed as \ref{eq:focal}.

\begin{equation}\label{eq:focal}
\begin{aligned}
FL(p_t) = -\alpha_t(1-p_t)^\gamma log(p_t)\\
p_t = \begin{cases} \mbox{p,} & \mbox{if } y=1 \\ \mbox{1-p,} & \mbox{otherwise} \end{cases}
\end{aligned}
\end{equation}

In the case of task B models, a Sparse Categorical Focal Loss was used, which is a focal loss implementation as an adapted multiclass softmax cross-entropy from the focal-loss package\footnote{\href{https://focal-loss.readthedocs.io/}{https://focal-loss.readthedocs.io/}}. After a few experiments with hyperparameters, they were set to \(\gamma=2, \alpha=1\). All models have been fine-tuned on available data with added development points, aside from 100 entries left out of the EDOS development dataset, which will be used for ensemble validation. Based on the results in Table \ref{table:4}, it can be observed that usage of Focal Loss provided performance benefits only in the case of a RoBERTa base model, while the rest degraded in performance. An interesting takeaway is the improvement of BERTweet from extended data, which in the case of RoBERTa large decreased the Macro-F1 score.

\begin{table}[]
\begin{tabular}{l|l|l|l|}
Model         & Dataset & Loss & Macro-F1          \\ \hline
RoBERTa base  & B1      & FL   & 0.680             \\ \hline
RoBERTa base  & B1      & CE   & 0.668             \\ \hline
BERTweet base & B1      & FL   & 0.671             \\ \hline
BERTweet base & B1      & CE   & 0.680             \\ \hline
RoBERTa large & B1      & FL   & 0.702             \\ \hline
RoBERTa large & B1      & CE   & \textbf{0.714}    \\ \hline
BERTweet base & B2      & CE   & 0.694             \\ \hline
RoBERTa large & B2      & CE   & 0.690   
\end{tabular}
\caption{Table depicting task B final model benchmark. The Macro-F1 values have been calculated on a validation subset of 100 points. FL - Focal Loss, CE - base Cross Entropy loss, dataset descriptions can be found in Section \ref{sec:eda}.}
\label{table:4}
\end{table}

Although the overall performance of models trained using the extended dataset B2 depicted by the Macro-F1 score does not show improvement, the insertion of EXIST data ameliorated the results on minority classes during testing. The most significant increase in data quantity by augmentation, described in Section \ref{sec:eda}, occurred for the `prejudiced discussions' class, and the performance reflects it. On average, between the same models fine-tuned on datasets B1 and B2, the F1 score for this class increased from 0.615 to 0.635 on the validation subset. The performance on the other altered class, `threats, plans to harm and incitement', which was injected with fewer entries, remained nearly identical between the two train sets.

\paragraph{Task C} For task C the idea of implementing Focal Loss due to task similarity was tested. In this case, all of the models actually degraded in performance; hence it was not used further. Due to previous testing, only specific models that performed well in the previous task were fine-tuned for each specific category. 

\subsection{Ensemble implementation}
\label{sec:ensemble}

\paragraph{Task A} The final step of system implementation consisted of creating an ensemble. For the first task of binary sexism detection, the final seven models presented in Table \ref{table:3} were chosen to be a part of the ensemble. As the set of 500 points was left out during training, it was used to test the performance. A soft voting ensemble has been created for all subsets from the set of seven final models that consist of two or more models. It is a popular method that averages the probability predicted by each model (\(p_{ij}\)). For M classifiers in a binary problem it can be formalized for the i-th entry as Equation \ref{eq:softvote} \cite{MOHAMMED20228825}.

\begin{equation}\label{eq:softvote}
\begin{aligned}
\hat{y} = \argmaxA_i \frac{1}{M} \sum_{j=1}^{M} p_{ij}
\end{aligned}
\end{equation}

The results of testing present in Table \ref{table:5} show that the best-performing ensemble for this validation set consists of five models - Toxicity RoBERTa, RoBERTa large, BERTweet base fine-tuned on dataset A2, and Toxicity RoBERTa, BERTweet base fine-tuned on dataset A3. This is an exciting observation that inserting models that used only SemEval data worsened the results on the validation set. Also, for each ensemble size the best score outperformed every single model on its own. Another interesting finding is that the second-best ensemble consists of 3 models: RoBERTa large fine-tuned on dataset A2, BERTweet base fine-tuned on the dataset A2, and Toxicity RoBERTa fine-tuned on dataset A3, which does not accommodate the best performing model---BERTweet base on dataset A3.

\begin{table}[]
\begin{tabular}{l|l|}
Ensemble size & Best Macro-F1          \\ \hline
1             & 0.865                  \\ \hline
2             & 0.871                  \\ \hline
3             & 0.885                  \\ \hline
4             & 0.877                  \\ \hline
5             & \textbf{0.891}         \\ \hline
6             & 0.879                  \\ \hline
7             & 0.879        
\end{tabular}
\caption{Task A results of the best ensembles for each size - a number of models used in the ensemble, 1 is a single model. Metrics calculated on a validation subset of 500 points.}
\label{table:5}
\end{table}

\begin{figure}[ht!]
\includegraphics[width=\columnwidth]{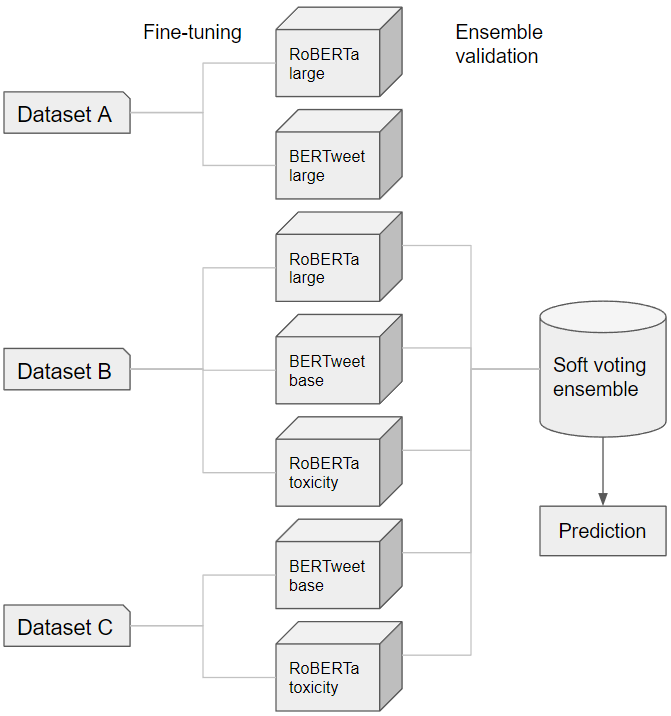}
\caption{Final ensemble creation process for task A}
\label{fig:ensemble}
\end{figure}

The selected five-model ensemble was used as the final solution to the problem in task A. The graphical overview of the process is represented by the graph in Figure \ref{fig:ensemble}. Further tests with an additional weighting of scores for each model were conducted. This led to improvements of around 0.2\% in terms of Macro-F1 but was not used as overfitting to the validation set might occur. 

\paragraph{Task B} The methodology for the final ensemble implementation in task B was similar to the first task. Five of the best models based on results in Table \ref{table:4} were selected: RoBERTa base, BERTweet base, and RoBERTa large on dataset B1, and RoBERTa large, BERTweet base on dataset B2. Only in the case of the RoBERTa base trained on EDOS data the version implemented with the Focal loss function was selected. The base Cross Entropy loss version was used in the remainder of the models. Task B presents a problem of multiclass classification problem; hence both soft voting and hard voting strategies were tested. Hard voting ensembles take the majority predicted class. In the case of this implementation, for an even number of labels, probabilities are taken into consideration, as in the soft voting equation \ref{eq:softvote}. The comparison between ensemble strategies was not as straightforward as tests conducted in task A, where a soft voting ensemble outperformed the hard voting method for every subset of models. For task B, an ensemble using all five models with a hard vote scored a Macro-F1 score of 0.711 compared to 0.702 for the soft strategy on the chosen validation set. The best solution overall, however, was a soft vote ensemble encompassing three models, two fine-tuned on dataset B1 containing only SemEval data: BERTweet base, RoBERTa large, and RoBERTa large on dataset B2 with EXIST data added. This solution scored a Macro-F1 of 0.724 on the validation set. Based on the results seen in Table \ref{table:6}, not all ensembles outperformed the single best model as was the case in task A and the enhancement of using this method is much less impactful than in task A.

\begin{table}[]
\begin{tabular}{l|l|}
Ensemble size & Best Macro-F1          \\ \hline
1             & 0.714                  \\ \hline
2             & 0.720                  \\ \hline
3             & \textbf{0.724}         \\ \hline
4             & 0.699                  \\ \hline
5             & 0.711              
\end{tabular}
\caption{Task B results of the best ensembles for each size - a number of models used in the ensemble, 1 is a single model. Metrics calculated on a validation subset of 100 points.}
\label{table:6}
\end{table}

\paragraph{Task C} The chosen architecture for tackling the problem in task C, based on previously described experiments, is first to use the ensemble implemented in the previous task to get category labels and then predict using different models for each label. This meant that for each of the four categories, a separate ensemble had to be created. With such an architecture, the number of used models grows quickly, so only three models were trained for each class---RoBERTa base, BERTweet base, and RoBERTa large, which amounts to a total of 12 models. Due to time and resource constraints, all three models were used in ensembles using soft voting techniques based on experiments conducted for task B. Due to the lack of test possibility on the validation set, it was included in the training data.

\section{Results}

\paragraph{Task A} The final solution was submitted for each subtask and ranked on the leaderboard. For the problem in task A, we achieved a Macro-F1 score of \textbf{0.838} and an accuracy of 0.880 with more detailed metrics depicted in Table \ref{table:7}, which placed us at the \textbf{36th} place out of 84 submissions. After the release of labelled test data, the remainder of the ensembles created in Section \ref{sec:ensemble} were tested. An ensemble that encompassed three models, the best ones on each dataset based on validation data results in Table \ref{table:3}: RoBERTa large on dataset A1, RoBERTa large on dataset A2, BERTweet base on dataset A3 achieved a Macro-F1 score of 0.858 on the test set, which far exceeds the performance of the submitted model. On the selected validation set, on the other hand, it scored 0.872 compared to 0.891 of the submitted system. These results show that the presented architecture is capable of achieving a much higher score but also that a method of ensemble creation from multiple varying models on different datasets is highly dependent on selected data, even if the class distribution is similar. The best performing single model was the BERTweet large fine-tuned on dataset A1 as it achieved a Macro-F1 score of 0.849 on the test set, and has been released on Hugging Face\footnote{\href{https://huggingface.co/tum-nlp/bertweet-sexism}{https://huggingface.co/tum-nlp/bertweet-sexism}}.

\begin{table}[]
\begin{tabular}{l|l|l|l|}
             & Precision & Recall & F1-score \\ \hline
Not sexist   & 0.92      & 0.92   & 0.92     \\
Sexist       & 0.75      & 0.76   & 0.76     \\ \hline
Accuracy     &           &        & 0.88     \\
Macro avg    & 0.84      & 0.84   & 0.84     \\
Weighted avg & 0.88      & 0.88   & 0.88    
\end{tabular}
\caption{Table showing main classification metrics of submitted solution for task A.}
\label{table:7}
\end{table}

\paragraph{Task B} The ensemble submitted to task B achieved a Macro-F1 score of \textbf{0.623}, which resulted in the \textbf{29th} place out of 69 submissions. Contrary to the results shown in Table \ref{table:4}, where most of the models performed similarly, on the test set RoBERTa large trained on dataset B1 scored a Macro-F1 of 0.642, while the rest of the models achieved 0.605 or worse. The best model remains the same on both datasets, but the discrepancy is much higher; hence, the selected ensemble worsened the result.

\paragraph{Task C} The performance of the final solution for task C was bound to the predictive capabilities of the previous model due to the architecture. Nevertheless, it managed to score a Macro-F1 of \textbf{0.457}, ranking us as the \textbf{25th} best team out of 63 submissions. In this case, the ensemble choice was fitting, outperforming the remainder of possible combinations while achieving better results than chosen models on their own.

\section{Conclusion}

In this paper, we describe a system for each of the three tracks for the EDOS task based on soft voting ensembles of multiple Transformer-based models. The solutions scored top 40\% on all three subtasks highlighting the consistency. We compare different methods of data augmentation and loss function adaptation to combat class imbalance. The impact of including additional balancing data from external sources on the performance of the models is explored. It is shown that increasing the number of minority class entries can increase efficiency in that class, but the overall performance might degrade. A comparison of performance on different test sets and the variation of ensemble results are discussed in detail, weighting its benefits and drawbacks. We also explore the benefits of using field-specific pre-trained models in comparison to the base counterparts. Plans for future work involve exploring and explaining the models to focus on the interpretability part of sexism detection, along with further comparisons between single-model architecture and the use of ensembles in that regard. Mitigating the negative impact of augmenting the dataset with minority class entries from external sources on majority class performance would be another development step in this study.

\section*{Limitations}

The final solution proposed by our team is based on ensembles. It has been shown in this work that they have the potential to increase the performance over a single model solution but come with some drawbacks. The resources required for training an ensemble consisting of five models, as in the solution for task A, are five times higher than a single model. Training time and computing power requirements are not the only constraints with such a solution, as the memory footprint is also increased. Designing a reliable ensemble can be difficult and require more testing, as using the best-performing models is often not the optimal solution which has been shown in previous work \cite{wenjia08}. The EDOS task focuses not only on the detection of online sexism but also on the explanation behind it in the form of hierarchical categorization. There are many other methods of enhancing interpretability. Some of them use specific features of a model, e.g. attention scores, to explore model behaviour, which would inherently be more difficult to extract from an ensemble \cite{bodria20}. 

\bibliography{anthology,custom}
\bibliographystyle{acl_natbib}

\end{document}